%% file: main.tex
\documentclass[12pt]{article}

\usepackage{algorithm}
\usepackage{algpseudocode}
\usepackage{amsmath}
\usepackage{amssymb}
\usepackage{array}
\usepackage[english]{babel}
\usepackage[T1]{fontenc}
\usepackage[margin = 25 mm]{geometry}
\usepackage{graphicx}
\usepackage{hyperref}
\usepackage{lmodern}
\usepackage{mathtools}
\usepackage{parskip}
\usepackage{pgffor}
\usepackage{xcolor}
\usepackage{empheq}
\usepackage{tikz,lipsum,lmodern}
\usepackage{tikz-cd} 
\usepackage[most]{tcolorbox}
\usepackage{enumitem}
\usepackage{footnote}
\makesavenoteenv{tabular}

\input{notation}

\definecolor{dkgreen}{rgb}{0,0.6,0}
\definecolor{gray}{rgb}{0.5,0.5,0.5}
\definecolor{mauve}{rgb}{0.58,0,0.82}

\title{Encoding architecture algebra}
\author{\{ Stephane Bersier\footnote{stephane.bersier@gmail.com}, \,Xinyi Chen-Lin\footnote{xinyitsenlin@gmail.com} \}}
\date{}

\begin{document}

\maketitle

\begin{abstract}
\input{sections/abstract}

\end{abstract}

\vspace{5mm}

\setcounter{tocdepth}{2}
\tableofcontents

\input{sections/introduction}

\input{sections/highLevel}

\input{sections/flatType}

\input{sections/tensor}

\input{sections/sumType}

\input{sections/productType}

\input{sections/multiset}

\input{sections/relationship}

\input{sections/additional}

\input{sections/example}

\input{sections/conclusion}

\appendix
\input{sections/appendix}

\bibliographystyle{IEEEtran}
\bibliography{ref}

\end{document}

%% file: notation.tex
\newcommand{\subtype}{\mathrel{\raisebox{0.08 em}{\scalebox{0.8}{<}}\mkern-1.5mu\scalebox{1.1}{$\vcentcolon$}}}

\newcommand{\Tensor}[1]{\texttt{Tens}[\, #1 \,]}
\newcommand{\Vector}[1]{\texttt{Vec}[\, #1 \,]}
\newcommand{\Yector}[1]{\texttt{Yec}[\, #1 \,]}
\newcommand{\Scalar}{\texttt{Scal}}
\newcommand{\Unit}{\texttt{Unit}}
\newcommand{\Prod}[1]{\texttt{Prod}[\, #1 \,]}
\newcommand{\Option}[1]{\texttt{Option}[\, #1 \,]}
\newcommand{\Enum}[1]{\texttt{Enum}[\, #1 \,]}
\newcommand{\Bool}{\texttt{Bool}}
\newcommand{\MSet}[1]{\texttt{MSet}[\, #1 \,]}
\newcommand{\Sum}[1]{\texttt{Sum}[\, #1 \,]}
\newcommand{\List}[1]{\texttt{List}[\, #1 \,]}

\newcommand{\map}{\texttt{map}}

%% file: sections/abstract.tex
Despite the wide variety of input types in machine learning, this diversity is often not fully reflected in their representations or model architectures, leading to inefficiencies throughout a model's lifecycle. This paper introduces an algebraic approach to constructing input-encoding architectures that properly account for the data's structure, providing a step toward achieving more typeful machine learning.

%% file: sections/introduction.tex
\section{Introduction}

There is growing awareness of the importance of designing model architectures that capture and respect the distinct structure of input data.
Many successful deep learning architectures, such as transformers \cite{vaswani2017attention}, convolutional neural networks (CNNs)\cite{lecun1998gradient}, graph neural networks (GNNs) \cite{scarselli2008graph}, and recurrent neural networks (RNNs)\cite{elman1990finding}, inherently incorporate aspects of data structure.

Ongoing research focuses on refining existing architectures, as well as designing new ones for other types of structured data. For instance, DeepSets \cite{zaheer2017deep} are tailored to process sets, group and gauge equivariant CNNs \cite{cohen2016group}\cite{cohen2019gauge} respect both global and local symmetries in the data, and strongly-typed RNNs \cite{balduzzi2016strongly} incorporate explicit types within recurrent networks. By accounting for the structure of the input data, these model architectures exhibit improved performance, better generalization with fewer parameters, and enhanced interpretability.

In parallel, the typeful programming paradigm \cite{cardelli1989typeful} has been gradually gaining traction in software engineering. Typeful programming is about faithfully and formally representing/modeling the mathematical structures that are relevant to the problem at hand. The machine learning (ML) community, however, remains largely type-unaware, reflected in how major ML frameworks mainly rely on a single, catch-all type known as ``tensor''— a Jack of all trades, but master of none.  

In this paper, we complement the existing work in structured machine learning by bringing more type-awareness into model design. Specifically, we propose a set of natural primitives to systematically build model architectures capable of handling arbitrarily complex algebraic data types (ADTs) as input. These primitives could be used as the theoretical basis for a higher-level typeful ML framework. 

The core idea is the generalization of dense linear layers, where the input can be of a composite type. We call these layers \emph{multilinear flattening layers} or MFLs.
Together with primitive operations and constructors for each type, they can be used to algebraically construct architectures operating on any ADT.

The paper is organized as follows. Section \ref{sec:highLevel} provides an overview of the high-level approach. In section \ref{sec:encodingType}, we introduce the flat type. The subsequent sections \ref{sec:tensors}, \ref{sec:sumType}, \ref{sec:productType} and \ref{sec:multiset} explore each ADT, discussing their constructors, primitive operations, MFLs, and relevant special cases. Section \ref{sec:typeRelations} examines relationships between some ADT cases. Additional considerations, such as architecture simplification and weight sharing, are covered in section \ref{sec:additional}. Examples are presented in section \ref{sec:example}, with the conclusion in section \ref{sec:conclusion}.

\subsection{Notation reference}
We use (multi)linear algebra notation, with bold lowercase letters representing vectors (e.g.\ $\boldsymbol{v}$), and bold uppercase letters representing other tensors or (multi)linear maps (e.g.\ $\boldsymbol{L}$). Components are denoted by the same letter in non-bold font, with subscripts indicating the axes (e.g.\ $v_i$ and $L_{ij}$).

\begin{tabular}{@{}ll}
  $\subtype$ & the (reflexive and transitive) subtype relation \\
  $\sigma$ & a permutation \\
  $l$ & a length variable (ranging over $\mathbb{N}$) \\
  $V$ & a vector space \\
  $T$ & a type variable \\
  $C$ & a type constructor variable \\
  $v_i$ & a vector component\\  
  $\boldsymbol{v}^{(r)}$ & a vector with a label $r$\\
  $\Unit$ & the unit type \\
  $\Scalar$ & the type of scalars\footnote{typically $\mathbb{R}$, often approximated in practice by floating-point numbers} \\
  $\Vector{l}$ & the type for vectors of dimensionality $l$ \\
  $\Yector{l}$ & the learned vector type of dimensionality $l$ \\
  $\Tensor{l_1, ..., l_n}$ & the tensor type with axes of length $l_1$ to $l_n$ \\
  $\Sum{T_1, ..., T_n}$ & the sum of types $T_1$ to $T_n$ \\
  $\Prod{T_1, ..., T_n}$ & the product of types $T_1$ to $T_n$ \\
  $(x_1, ..., x_n)$ & a tuple\footnote{a.k.a. a product type value} containing values $x_1$ to $x_n$ \\
  $\MSet{T}$ & the type of multisets with elements of type $T$\\
  $\Option{T}$ & the type of optional values of type $T$\\
  $T \rightarrow T'$ & the function type from $T$ to $T'$ \\
  $(x_1, ..., x_n) \rightarrow \emph{<body>}$ & a function value\footnote{a.k.a.\ lambda, anonymous function, function literal} that maps $(x_1, ..., x_n)$ to \emph{<body>} \\  
\end{tabular}

%% file: sections/highLevel.tex
\section{Approach}\label{sec:highLevel}

\subsection{Algebraic data types (ADTs)}

The most basic flavor of ADTs includes only combinations of sums and products, as well as the unit type as the base case. However, the idea of algebraic composition can be extended to more base and inductive cases.

In this paper, we consider a richer base case, namely tensors, of which vectors, scalars and unit types are special cases, and one additional inductive case, namely multisets. The ADTs we consider are thus of one of the four forms below:
\begin{itemize}
    \item $\Tensor{l_1, ..., l_n}$
    \item $\Sum{T_1, ..., T_n} $
    \item $\Prod{T_1, ..., T_n}$
    \item $\MSet{T}$
\end{itemize}
where $T$ and $T_i$ with $i=1,..., n$ are themselves ADTs. 

Each ADT case comes with a set of allowed operations. Additionally, each inductive case has its own value constructors, and type constructor. The latter is covariant in its type parameters:
\begin{equation}
    C[\,T_1, ..., T_n\, ] <: C[\,T'_1, ..., T'_n\, ]
    \quad 
    \text{where:}  \quad T_i <: T'_i, \quad i=1,..., n
\end{equation}

\subsection{Architecture algebra}
The architecture algebra for the encoding of inputs is based on the following groups of primitives:
\begin{itemize}
    \item Multilinear flattening layers (MFLs)
    \item Poly mapping\footnote{See subsection \ref{sec:polymap}.}
    \item Fixed operations for each type
    \item Constructors for the inductive types
\end{itemize}
Primitives can be composed (in accordance with their type signatures) to construct model architectures for the encoding of ADTs.

Though we will not explore this further here, the architecture algebra can also be framed categorically. In this view, the objects of the category are ADTs, with type constructors acting as endofunctors. The morphisms of the category are architectures, composed of primitives (or only the MFLs, for a more restricted category), converting the source type to the target type.

\subsection{Multilinear flattening layers (MFLs)}
MFLs can be viewed as generalizations of dense (a.k.a.\ fully connected) linear layers. Whereas dense linear layers map from vector to coordinate vector, leading to a trivial type signature $\Vector{l'} \rightarrow \Yector{l}$, MFLs can handle non-trivial input types. They are defined for four basic input types, with corresponding type signatures:
\begin{itemize}
    \item $\Tensor{l_1, ..., l_n} \rightarrow \Yector{l}$
    \item $\Sum{\Vector{l_1}, \,...,\, \Vector{l_n}} \rightarrow \Yector{l}$
    \item $\Prod{\Vector{l_1}, \,...,\, \Vector{l_n}}\rightarrow \Yector{l}$
    \item $\MSet{\Vector{l'}}\rightarrow \Yector{l}$
\end{itemize}
where the flat type $\Yector{l}$ is a subtype of the vector type $\Vector{l}$, see section \ref{sec:encodingType}.

To find the right MFL architecture for each case, the following (possibly redundant) properties were used as guiding principles:
\begin{itemize}
    \item Minimality
    \item Invariance under changes in input vector basis
    \item Invariance under changes in output vector basis
    \item Invariance under input type isomorphisms
    \item No invariance under other input transformations
    \item Multilinearity (when the input is bias-augmented, see \ref{sec:linearAffine})
    \item Image that spans the output space
\end{itemize}

\subsection{Poly mapping} \label{sec:polymap}
We extend the polytypic map function to accept multiple functions as arguments, a generalization we refer to as \emph{poly mapping}. Its type signature is:
\begin{equation}
    \map:\;\Prod{T_1\rightarrow T'_1,\,..., \,T_n\rightarrow T'_n,  \, C[\,T_1, ..., T_n\,]} \rightarrow C[\,T'_1, ..., T'_n\,]
\end{equation}
where $C$ denotes any composite type\footnote{In the (tensor type) base case, where there are no type arguments, the polytypic map is trivial, taking no function arguments.}. $\map$ is derivable from the primitive operations and constructors specific to each ADT. It plays a crucial role in enabling transformations of constituent types within composite types.

\subsection{Linear versus affine}\label{sec:linearAffine}
Dense linear layers are commonly written in two algebraically equivalent forms. The affine form explicitly shows the bias term, while the linear form absorbs the bias by \emph{bias-augmenting} the input vector: 
\begin{equation}\label{eq:denseLayer}
    \boldsymbol{y} = \boldsymbol{b} + \boldsymbol{L} \,\boldsymbol{v} 
    \quad  \longleftrightarrow \quad 
    \boldsymbol{y} =  \boldsymbol{L}' \,\boldsymbol{v}', \quad    
    v'_i =
    \begin{cases}
        & 1 \quad \; \operatorname{if} \quad i= 0    \\
        &v_{i} \quad \operatorname{if} \quad i=1, ..., l
    \end{cases}
\end{equation}
where $\boldsymbol{v}$ and $\boldsymbol{y}$ are the input and output vectors, respectively, $\boldsymbol{v}'$ is the bias-augmented input vector, and $\boldsymbol{L}'$ is the augmented weight matrix that contains the bias vector $\boldsymbol{b}$ and the weight matrix $\boldsymbol{L}$.

This linear/affine correspondence naturally generalizes to a multilinear/multiaffine relationship, which is used in our approach to deriving MFLs. Specifically, we first derive the MFLs in multilinear form, then obtain the multiaffine form by explicitly setting the bias components of the bias-augmented inputs to 1, and merging all the resulting bias terms that are equivalent.

%% file: sections/flatType.tex
\section{The flat type} \label{sec:encodingType}
A vector\footnote{The vector type is discussed as a special case of the tensor type, see \ref{sec:vectors}} should not be confused with its coordinate representation. The two are in bijection only when a basis is fixed.

In machine learning, a learned vector comes with an implicit learned basis, providing meaningful interpretation to its components. Thus, a learned vector can be understood as having a more refined type with additional operations. We refer to such vectors as \emph{yectors} when represented in their learned basis:
\begin{equation}
    \Yector{l} \subtype \Vector{l}
\end{equation}
Each MFL maps a composite type to a yector type, which is the flat type. 

\subsection{Operations}
\begin{tcolorbox}[left=1pt, colback=white, colframe=black, title=Yector]
\begin{align*}
    \texttt{map}&:\; \Prod{\Scalar \rightarrow \Scalar, \, \Yector{l}} \rightarrow \Yector{l}\\[0.4em]
    & \boldsymbol{v}' = \texttt{map}(f, \boldsymbol{v}), \quad v'_k = f(v_k)\\
    &\\
     \boldsymbol{\cdot} \; &:\;\Prod{\Yector{l} , \Yector{l}} \rightarrow \Scalar  &(\text{dot product})\\[0.4em]
    & s  = \boldsymbol{v} \cdot \boldsymbol{w} = \sum_{k} \,  v_k w_k    \\
    &\\[-0.75em]
    + \;  &:\; \Prod{\Yector{l} , \Yector{l}} \rightarrow \Yector{l}&(\text{addition}) \\[0.4em]
    & \boldsymbol{y} = \boldsymbol{v} + \boldsymbol{w}, \quad y_k = v_k + w_k   \\
    &
\end{align*}
\end{tcolorbox}

The map\footnote{This is not a poly map, it maps a function over coordinates.} operation applies functions component-wise. When the applied function $f$ is nonlinear, it can be thought of as an activation function. Of course, dense linear layers \eqref{eq:encodingVector} can be interleaved with such activation layers to add depth to any architecture.

The addition operation and the dot product (a special case of the tensor contraction) are inherited from tensor operations in \ref{sec:tensorOps}.

%% file: sections/tensor.tex
\section{Tensor types}\label{sec:tensors}

Tensors\footnote{Note that these mathematical tensors are to be distinguished from what frameworks like TensorFlow and PyTorch call tensors, which are more accurately called multidimensional arrays. } are elements of the tensor product of vector spaces\footnote{We generalize the traditional definition of tensors that live in the tensor product of the same vector space and its dual.}:
\begin{equation}\label{eq:tensorProductVs}
    V^{(1)} \otimes \cdots \otimes V^{(n)}
\end{equation}
where the vector spaces $V^{(r)}, r=1,..., n$ are over the real numbers\footnote{Everything should also apply to vector spaces over the complex numbers.} and have dimensionality $l_r \in \mathbb{N}$. 

Just as vectors, tensors are geometric invariants which can be represented by components in different bases. For instance:
\begin{align}
    \boldsymbol{N} = \sum_{i_1, ..., i_n} N_{i_1 ... i_n} \mathbf{e}^{(1)}_{i_1}\otimes \cdots \otimes \mathbf{e}^{(n)}_{i_n}
    = \sum_{i_1, ..., i_n} N'_{i_1 ... i_n} \mathbf{e}'^{(1)}_{i_1}\otimes \cdots \otimes \mathbf{e}'^{(n)}_{i_n} 
\end{align}
where $\{\mathbf{e}^{(r)}_{i_r}\}$ and $\{\mathbf{e}'^{(r)}_{i_r}\}$ with $i_r=1,..., l_r$ are two distinct bases that span $V^{(r)}$.

The corresponding tensor types do not depend on any other algebraic type and are fully parameterized by axis lengths (a.k.a.\ tensor shape):
\begin{equation}
    \Tensor{l_1, ..., l_n}, \quad l_i \in \mathbb{N}, \quad  i=1,..., n
\end{equation}
The number of axes $n$ is known as the order (or rank) of the tensor.

\subsection{Isomorphisms}

The tensor type has the following isomorphisms:
\begin{itemize}
    \item Axis reordering: $\Tensor{l_1,..., l_n} \cong \Tensor{l_{\sigma(1)},..., l_{\sigma(n)}}$
    \item Scalar collapse: $\Tensor{l_1,..., l_n,1 } \cong \Tensor{l_1,..., l_n}$
    \item Unit collapse: $\Tensor{l_1,..., l_n, 0} \cong \Tensor{0}$
\end{itemize}

\subsection{Operations}\label{sec:tensorOps}
\begin{tcolorbox}[left=1pt, colback=white, colframe=black, title=Tensor]
\begin{align*}
    \otimes \; &:\; \Prod{\Tensor{l_1, ..., l_m}, \, \Tensor{l'_1, ..., l'_n}} \\
    & \qquad \qquad \rightarrow \Tensor{l_1, ..., l_m, \,l'_1, ..., l'_{n}} &(\text{tensor product})\\[0.4em]
    & \boldsymbol{N} = \boldsymbol{P} \otimes \boldsymbol{Q}, \quad 
    N_{i_1\cdots i_m \,j_1 \cdots j_n}  =  P_{i_1\cdots i_m} Q_{j_1\cdots j_n}  \\
    &\\
    \left<\;,\;\right>&:\; \Prod{\Tensor{l_p,l_q},\, \Tensor{l_1, ..., l_p, ..., l_q, ... l_{n}} }\\
    &\qquad \qquad \rightarrow \Tensor{l_1, ... , l_{p-1}, l_{p+1}, ..., l_{q-1}, l_{q+1}, ..., l_{n}}  &(\text{contraction})\\[0.4em]
    & \boldsymbol{N}' = \left<\boldsymbol{G},\, \boldsymbol{N}\right>, 
    \quad  N'_{i_1 \cdots i_{p-1} i_{p+1} \cdots i_{q-1} i_{q+1} \cdots i_n }  = \sum_{i_p, i_q} G_{i_p i_q} \,  N_{i_1\cdots i_p\cdots i_q\cdots i_n}     \\
    &\\[-0.75 em]
    +  \; &:\; \Prod{\Tensor{l_1, ..., l_n}, \, \Tensor{l_1, ..., l_n}} \rightarrow \Tensor{l_1, ..., l_n} &(\text{addition}) \\[0.4em]
    & \boldsymbol{N} = \boldsymbol{P} + \boldsymbol{Q}, \quad  N_{i_1 \cdots i_n }  = P_{i_1\cdots i_n} + Q_{i_1\cdots i_n}   \\
    &
\end{align*}
\end{tcolorbox}
Note that since our tensors reside in distinct vector spaces, see \eqref{eq:tensorProductVs}, a pairing bilinear map $\boldsymbol{G}$ is required to generalize the standard tensor contraction.

Rescaling is a special case of the tensor product when one of the arguments is a scalar.

\subsection{MFL}\label{sec:tensorMFL}

First, let us augment the tensor by adding bias dimensions to each of its axes. The type signature of the bias-augmentation operation is:
\begin{equation}
    \Tensor{l_1, ..., l_n}  \rightarrow \Tensor{l_1+1, \,..., \, l_n+1}
\end{equation}
and the augmented tensor is defined as follows: 
\begin{equation}
    N'_{i_1\cdots i_n} = 
    \begin{cases}
        & 1 \quad \operatorname{if} \quad \; \exists\,r \; \text{ s.t. } \; i_r = 0 \\
        &N_{i_1\cdots i_n} \quad \operatorname{otherwise}
    \end{cases}
\end{equation}

\begin{figure}
    \centering
    \includegraphics[width=\linewidth]{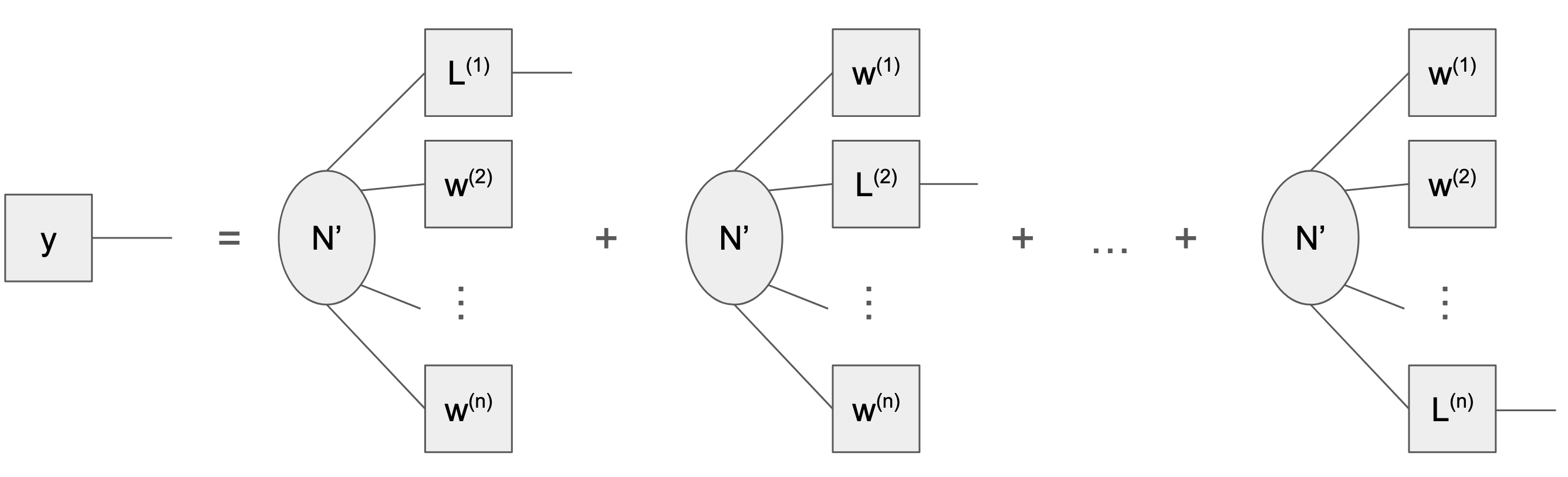}
    \caption{The tensor network diagram for the tensor MFL \eqref{eq:augmentedTensorFlattening}. Shapes are tensors, lines represent indices, and connected lines are contracted indices.}
    \label{fig:tensorFlattening}
\end{figure}

The flattening for the bias-augmented tensor type, with signature:
\begin{equation}
    \Tensor{l_1+1, \,...,\, l_n+1} \rightarrow \Yector{l}
\end{equation}
is defined by an architecture that must correctly handle tensor structure, using the previously defined tensor operations:
\begin{equation} \label{eq:augmentedTensorFlattening}
    y_k = \sum_{i_1 = 0}^{l_1} \cdots \sum_{i_n = 0}^{l_n}  \left( L^{(1)}_{k \,i_1} w^{(2)}_{i_2} \cdots w^{(n)}_{i_{n}} 
    +\cdots 
    +w^{(1)}_{i_1} \cdots w^{(n-1)}_{i_{n-1}} L^{(n)}_{k \,i_n} \right)
    N'_{i_1 ... i_n}, \quad k=1,..., l
\end{equation}
where $\boldsymbol{w}^{(r)}$ and $\boldsymbol{L}^{(r)}$, for $r=1,...,n$, are weight vectors and matrices, respectively\footnote{The pairing bilinear maps in tensor contractions are absorbed into the weights.}. Figure~\ref{fig:tensorFlattening} shows the corresponding tensor network diagram.

The MFL architecture for the original tensor (i.e.\ without bias augmentation), with type signature:
\begin{equation}
    \Tensor{l_1, ..., l_n} \rightarrow \Yector{l}
\end{equation}
is obtained from equation \eqref{eq:augmentedTensorFlattening} by explicitly setting all the bias components to 1:
\begin{equation}\label{eq:encodingTensor}
    y_k = b_k + \sum_{j_1 = 1}^{l_1}  \cdots \sum_{j_n = 1}^{l_n}  \left(L^{(1)}_{k \,j_1} w^{(2)}_{j_2} \cdots w^{(n)}_{j_{n}} +\cdots 
    +w^{(1)}_{j_1} \cdots w^{(n-1)}_{j_{n-1}} L^{(n)}_{k \,j_n} \right) N_{j_1... j_n}
\end{equation}
where all the terms involving the bias dimensions simplify to one single bias vector, see \eqref{eq:multiSumWithBias}. The total number of weights is then:
\begin{equation}
    l+(1+l) \sum_{i=1}^n l_i 
\end{equation}

\subsection{Special cases}
\subsubsection{Vector types}\label{sec:vectors}
Vector types correspond to tensor types with one axis:
\begin{equation}
    \Vector{l} = \Tensor{l}
\end{equation}
As a result, the MFL architecture \eqref{eq:encodingTensor} reduces to the standard densely connected linear layer, as it should:
\begin{equation}\label{eq:encodingVector}
    \boldsymbol{y} = \boldsymbol{b} + \boldsymbol{L}\,\boldsymbol{v}
\end{equation}

\subsubsection{Scalar type}\label{sec:scalarTypes}
The scalar type is equivalent to the vector type with one dimension:
\begin{equation}
    \Scalar = \Vector{1}
\end{equation}

\subsubsection{Unit type}\label{sec:unitTypes}
The unit type is equivalent to the extremal case of the vector type with no dimensions:
\begin{equation} 
    \Unit = \Vector{0} 
\end{equation}
Its only value is the empty vector. 
The linear layer \eqref{eq:encodingVector} degenerates to a single bias vector, also known as an embedding:
\begin{equation}\label{eq:encodingUnit}
    \boldsymbol{y}=\boldsymbol{b}
\end{equation}

%% file: sections/sumType.tex
\section{Sum types}\label{sec:sumType}

A sum type, also known as a tagged union type, is a data type that can hold values from one of several predefined types, but only one at a time. Each possible type is tagged to distinguish which variant is currently being held. Sum types are useful for representing data that can take on multiple shapes.

\subsection{Constructors}
The constructor of a sum type is specialized by an index representing the type tag:
\begin{equation}
\texttt{case}_{[T_1,..., T_n]}(i)(t_i) \quad \text{is of type} \quad \Sum{T_1,..., T_n}
\end{equation}
where $t_i$ is of type $T_i$.

\subsection{Isomorphisms}
Sum types are isomorphic under argument reordering: 
\begin{equation}
    \Sum{T_1,..., T_n} \cong \Sum{T_{\sigma(1)},..., T_{\sigma(n)}}
\end{equation}
and a sum type with a single argument is isomorphic to it:
\begin{equation}
    \Sum{T} \cong T
\end{equation}

\subsection{Subtyping relations}
A sum type is covariant in its type parameters:
\begin{equation}
    \Sum{T_1, ..., T_n} <: \Sum{T'_1, ..., T'_n}
    \quad 
    \text{where:}  \quad T_i <: T'_i, \quad i=1,..., n
\end{equation}
and
\begin{equation}
    \Sum{T_1, ..., T_{n-1}} <: \Sum{T_1, ..., T_{n-1}, T_n}
\end{equation}

\subsection{Operations}
\begin{tcolorbox}[colback=white, colframe=black, title=Sum]
\begin{equation*}
    \texttt{analyze}: \,\Prod{T_1\rightarrow T,..., T_n\rightarrow T,  \, \Sum{T_1, ... , T_n}} \rightarrow T
\end{equation*}
\end{tcolorbox}
The analyze operation allows for case analysis (a.k.a.\ pattern matching).

The poly map operation in \ref{sec:polymap} can be implemented using \texttt{analyze} and the constructor(s):
\begin{align*}
    \texttt{map}&:\;   \Prod{T_1\rightarrow T'_1, \,...,\, T_n\rightarrow T'_n,  \, \Sum{T_1, ..., T_n}} \rightarrow \Sum{T'_1, ..., T'_n}\\
    &\texttt{map}(f_1, ..., f_n, s) = \texttt{analyze}(\texttt{case}_{[T_1, ..., T_n]}(1) \circ f_1, \,..., \,\texttt{case}_{[T_1, ..., T_n]}(n) \circ f_n, \,s)
\end{align*}

\subsection{MFL}
The flattening:
\begin{equation}
    \Sum{ \Vector{l_1}, \,...,\, \Vector{l_n}} \rightarrow \Yector{l} 
\end{equation}

is achieved using case analysis, where a separate linear layer \eqref{eq:encodingVector}:
\begin{equation}
    \Vector{l_r} \rightarrow \Yector{l},  \quad  r=1,..., n
\end{equation}
is applied to each case, that is:
\begin{equation}
    \boldsymbol{y} = 
    \begin{cases}
        \;\boldsymbol{b}^{(1)} + \boldsymbol{L}^{(1)} \boldsymbol{v}^{(1)} \\
        \;\qquad \vdots \\
        \;\boldsymbol{b}^{(n)} + \boldsymbol{L}^{(n)} \boldsymbol{v}^{(n)} 
    \end{cases}  
\end{equation}

\subsection{Union types}
Union types can be reduced to sum types by first converting the types in the union to a disjoint set of types.

\subsection{Special cases}
\subsubsection{Enums}
The enumeration types or enums correspond to extremal sum types, where each argument of the sum is the unit type, see \ref{sec:unitTypes}:
\begin{equation}
    \Enum{l} = \Sum{\Unit,\, ...,\, \Unit}
\end{equation}
where $l$ is the number of arguments in the sum type.

\subsubsection{Boolean}
The boolean type is equivalent to: 
\begin{equation}
    \Bool = \Enum{2} = \Sum{\Unit,\, \Unit}
\end{equation}

\subsubsection{The empty type}

While of little direct practical interest, the empty sum type is equivalent to the uninhabited type:
\begin{equation}
    \texttt{Nothing} = \Sum{}
\end{equation}

\subsubsection{Option types}
Option types (a.k.a.\ maybe types) are equivalent to sum types with two cases. An option type value is either empty, or it is a wrapped value from the underlying type:
\begin{equation}
    \Option{T}=\Sum{\Unit,\,T}
\end{equation}

%% file: sections/productType.tex
\section{Product types}\label{sec:productType}

A product type, whose values are often referred to as a tuples, is a data type that groups multiple values, each from potentially different types, into a single composite structure. Each element of a tuple can be accessed individually, making product types useful for representing data with multiple fields.

\subsection{Constructors}
We use tuple notation for product type literals:
\begin{equation}
    (t_1, ..., t_n) \quad \text{is of type} \quad \Prod{T_1, ..., T_n}
\end{equation}

\subsection{Isomorphisms}
Product types are isomorphic under argument reordering: 
\begin{equation}
    \Prod{T_1,..., T_n} \cong \Prod{T_{\sigma(1)},..., T_{\sigma(n)}}
\end{equation}
and a product type with a single argument is isomorphic to it:
\begin{equation}
    \Prod{T} \cong T
\end{equation}

\subsection{Subtyping relations}
A product type is covariant in its type parameters:
\begin{equation}
    \Prod{T_1, ..., T_n} <: \Prod{T'_1, ..., T'_n}
    \quad 
    \text{where:}  \quad T_i <: T'_i, \quad i=1,..., n
\end{equation}
and
\begin{equation}
    \Prod{T_1, ..., T_{n-1}, T_n} <: \Prod{T_1, ..., T_{n-1}}
\end{equation}

\subsection{Operations}
\begin{tcolorbox}[colback=white, colframe=black, title=Product]
\begin{equation*}
    \texttt{project}:\,\Prod{i \in [1,..., n],\, \Prod{T_1, ... , T_n}} \rightarrow T_i
\end{equation*}
\end{tcolorbox}

The poly map operation in \ref{sec:polymap} can be implemented using \texttt{project} and the constructor:
\begin{align*}
    \texttt{map}&:\;   \Prod{T_1\rightarrow T'_1,..., T_n\rightarrow T'_n,  \, \Prod{T_1, ..., T_n}} \rightarrow \Prod{T'_1, ..., T'_n}\\
    &\texttt{map}(f_1, ..., f_n, \,p) 
    = \, \left(\,f_1(\hspace{0.05em}\texttt{project}(1, p)\hspace{0.05em}),\,...,\, f_n(\hspace{0.05em}\texttt{project}(n, p)\hspace{0.05em}
)\,\right)
\end{align*}

\subsection{MFL}
Consider a product type of vectors. As with the tensor case, we begin by bias-augmenting the vectors. The type
signature of the bias-augmentation operation is:
\begin{equation}
    \Prod{\Vector{l_1},\, ...,\, \Vector{l_n}} \rightarrow \Prod{\Vector{l_1+1},\, ...,\, \Vector{l_n+1}}
\end{equation}
and the bias-augmented vectors are:
\begin{equation}
    v'^{(r)}_i =
    \begin{cases}
        & 1 \qquad \operatorname{if} \quad i = 0 \\
        &v^{(r)}_i \quad \operatorname{otherwise}
    \end{cases}
    \llap{} \quad, \quad \text{for} \quad r = 1, ..., n
\end{equation}

\begin{figure}
    \centering
    \includegraphics[width=0.5\linewidth]{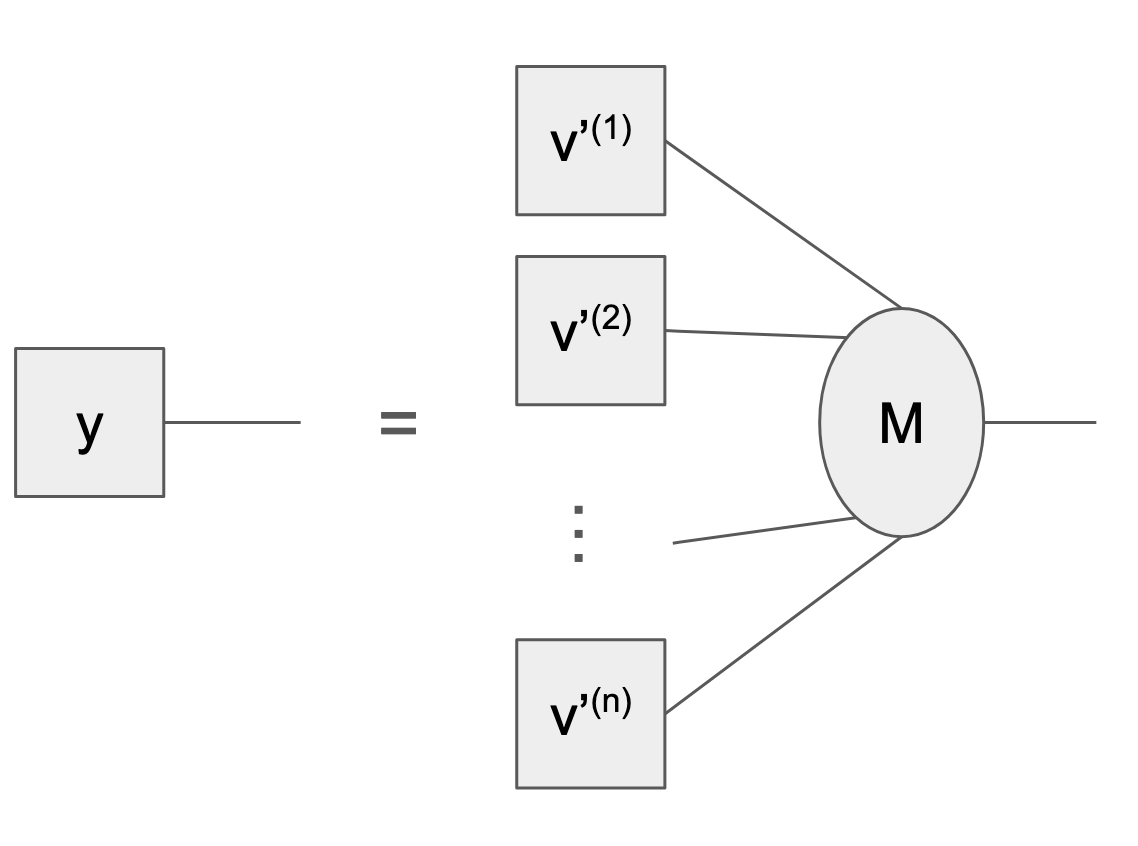}
    \caption{The tensor network diagram for the product type MFL \eqref{eq:encodingProductMultilinear}. Shapes are tensors, lines represent indices, and connected lines are contracted indices. }
    \label{fig:prodFlattening}
\end{figure}

The flattening for the product type of bias-augmented vectors, with type signature:
\begin{equation}
    \Prod{\Vector{l_1+1}, \,...,\, \Vector{l_n+1}} \rightarrow \Yector{l}
\end{equation}
is achieved by projecting all components out, then combining them with the tensor operations in \ref{sec:tensorOps}:
\begin{align} \label{eq:encodingProductMultilinear}
    y_k &= \sum_{i_1=0}^{l_1}\cdots \sum_{i_n=0}^{l_n} M_{k\, i_1\cdots i_n } \, v'^{(1)}_{i_1} \cdots v'^{(n)}_{i_n}, \quad k=1,...,l.
\end{align}
The corresponding tensor network diagram is shown in fig.~\ref{fig:prodFlattening}.

The MFL architecture in terms of the original vectors (i.e.\ not bias-augmented), with type signature:
\begin{equation}
    \Prod{\Vector{l_1}, \,...,\, \Vector{l_n}} \rightarrow \Yector{l}
\end{equation}
is obtained from
\eqref{eq:encodingProductMultilinear} by explicitly setting the bias components to one. Then, the sum decomposes into a sum of terms that correspond to different degrees of interaction between the vectors, see also \eqref{eq:encodingProductExpanded}:
\begin{align} \label{eq:encodingProduct}
    y_k &= b_k + \sum_{r} \sum_{j_r} L_{k\, j_r }^{(r)} \, v_{j_r}^{(r)} 
    + \sum_{r<s} \sum_{j_r,j_s} C_{k\, j_r j_s }^{(r,s)} \, v_{j_r}^{(r)} v_{j_s}^{(s)}  \nonumber \\
    &+ \sum_{r<s<t} \sum_{j_r,j_s, j_t} C_{k\, j_r j_s j_t}^{(r,s, t)} \, v_{j_r}^{(r)} v_{j_s}^{(s)}v_{j_t}^{(t)}  +\cdots + \sum_{j_1,..., j_n} C_{k\, j_1\cdots j_n}^{(1,..., n)} \, v_{j_1}^{(1)} \cdots v_{i_n}^{(n)} 
\end{align}
where we renamed the weight tensor components:
\begin{align*}
    b_k &= M_{ k\, 0\cdots 0 } \\
    L^{(r)}_{k\, j_r} &= M_{ k\, 0\cdots 0 \,j_r\,  0\cdots 0 } \\  
    C^{(r,s)}_{k\, j_r j_s} &= M_{ k\, 0\cdots 0\, j_r \, 0\cdots 0 \,j_s\,  0\cdots 0 }\\
    &\vdots\\
    C_{k\, j_1\cdots j_n}^{(1,..., n)} &= M_{k\, j_1\cdots j_n}
\end{align*}

Equation \eqref{eq:encodingProduct} can be written more neatly in multilinear algebra notation:
\begin{align}\label{eq:productMFL_matrixForm}
   \boldsymbol{y}  &= \boldsymbol{b} + \sum_{r}  \boldsymbol{L}^{(r)} \boldsymbol{v}^{(r)} 
    + \sum_{r<s}   \boldsymbol{C}^{(r,s)}( \boldsymbol{v}^{(r)}, \boldsymbol{v}^{(s)})  \nonumber\\
    &+ \sum_{r<s<t}  \boldsymbol{C}^{(r,s, t)} (\boldsymbol{v}^{(r)}, \boldsymbol{v}^{(s)},\boldsymbol{v}^{(t)}) +\cdots + \boldsymbol{C}^{(1,..., n)} (\boldsymbol{v}^{(1)}, ..., \boldsymbol{v}^{(n)}).
\end{align}

Note that the first-order term is equivalent to applying a linear layer to the concatenated feature vectors. The first non-linear interaction is the quadratic term, which takes pairwise interactions into account. The $r$th-order term takes into account $r$-range interactions. In practice, one may neglect higher-order interactions by truncating the sum above, as the number of weights in the terms grows combinatorially.

\subsection{Special case}
The unit type can also be seen as the extremal case of the empty product type. The MFL \eqref{eq:productMFL_matrixForm} reduces to an embedding \eqref{eq:encodingUnit}.

%% file: sections/multiset.tex
\section{Multiset types}\label{sec:multiset}
A multiset is like a set, except that duplicates are allowed. Multisets are useful for grouping values of a shared type when order is irrelevant but frequency matters, and also when the number of values to be grouped might vary.

\subsection{Constructor}
We use curly bracket notation to represent multiset literals:
\begin{equation}
    \{x_1, ..., x_n\}, \quad \forall \; n \in \mathbb{N} \quad \text{is of type} \quad \MSet{T}
\end{equation}

\subsection{Subtyping relation}
A multiset type is covariant in its type parameter:
\begin{equation}
    \MSet{T} <: \MSet{T'}
    \quad 
    \text{where}  \quad T <: T'
\end{equation}

\subsection{Operations}
\begin{tcolorbox}[colback=white, colframe=black, title=Multiset]
 \begin{align*} 
    \texttt{fold}&: \,\Prod{T',\, \Prod{T',\,T}\rightarrow T', \, \MSet{T} }\rightarrow T'\\ 
    &\\
    \cup \; &:\, \Prod{\MSet{T}, \,\MSet{T}}\rightarrow \MSet{T} &(\text{union})\\
 \end{align*}
\end{tcolorbox}

An additional constraint on the usage of the fold operation (that is not reperesented in its type signature above) is that it must take a function $f: \, \Prod{T',\,T}\rightarrow T'$ that satisfies a commutativity condition on its second argument:
\begin{equation}
    f(f(a, x_1),x_2) = f(f(a, x_2),x_1)
\end{equation}

Using the primitive operations and the constructor, it is possible to determine the multiplicity of a multiset element, add elements to a multiset, and derive:
\begin{align*}
    +\, &:\, \Prod{\MSet{T} , \,\MSet{T}}\rightarrow \MSet{T} & (\text{sum})\\ 
    &\\
    \cap \, &:\, \Prod{\MSet{T} , \,\MSet{T}}\rightarrow \MSet{T} & (\text{intersection})\\ 
    &\\
    -\; &:\, \Prod{\MSet{T} , \,\MSet{T}}\rightarrow \MSet{T} & (\text{difference})
\end{align*}

The poly mapping in \ref{sec:polymap} can be implemented as follows:
\begin{align*}
    \texttt{map}&:\Prod{T\rightarrow T', \, \MSet{T}} \rightarrow \MSet{T'}\\
    &\texttt{map}(f, s) = \texttt{fold}(\{\}, (a, x)\rightarrow {a \cup \{f(x)\}}, \,s)
\end{align*}

The cartesian product and some other operations are listed and implemented below: 
 \begin{align*} 
    \times \, &:\,\Prod{\MSet{T} , \,\MSet{T'}}\rightarrow \MSet{\Prod{T, \,T'}}\\
    &s_1 \times s_2 = \texttt{flatten}(\,\texttt{map}(\,x_1 \rightarrow\texttt{map}(\, x_2 \rightarrow (x_1, \, x_2), \; s_2\,),\, s_1\,)\,)\\
    &\\
    \texttt{flatten} &:\,\MSet{\MSet{T}} \rightarrow \MSet{T} \\
    & \texttt{flatten}(s) = \texttt{fold}(\, \{\}, (c, \,x) \rightarrow c \cup x, \,s \,)\\ 
    &\\    
    \texttt{reduce}&:\,\Prod{\MSet{T},\, M} \rightarrow T, \quad \text{where $M$ is a commutative monoid $(+, e)$ on $T$}\\
    &\texttt{reduce}(s, m) =\texttt{fold}(\,e,\; (c, \,x) \rightarrow c + x, \; s\,)\\
    &\\
    \texttt{size}&:\, \MSet{ T} \rightarrow \mathbb{N}\\
    &\texttt{size}(s) =\texttt{reduce}(\,\map( x \rightarrow 1, s),\, (0, +)\,)\\
 \end{align*}

\subsection{MFL}
The type signature of the multiset MFL is:
\begin{equation}
    \MSet{\Vector{l'}} \rightarrow \Yector{l}
\end{equation}
The architecture is obtained by folding the elements via summation, see Theorem 2 in \cite{zaheer2017deep}.
For a given multiset:
\begin{equation}
    \{\boldsymbol{v}^{(1)}, \,...,\, \boldsymbol{v}^{(n)}\}
\end{equation}
the multiset MFL in linear and affine forms are
\begin{equation}\label{eq:encodingMSet}
    \boldsymbol{y} = \sum_{r=1}^n  \boldsymbol{L}' \, \boldsymbol{v}'^{(r)} = n \,\boldsymbol{b} + \sum_{r=1}^n  \boldsymbol{L} \, \boldsymbol{v}^{(r)}
\end{equation}
where $\boldsymbol{b}$ is the bias vector,  $\boldsymbol{L}$ is the weight matrix and the apostrophes here indicate bias-augmentation, see \ref{sec:linearAffine}. 

In practice, a normalized form of \eqref{eq:encodingMSet}, i.e. 
\begin{equation}
    \boldsymbol{y'} = \,\boldsymbol{b} + \frac{1}{n}\sum_{r=1}^n  \boldsymbol{L} \, \boldsymbol{v}^{(r)}
\end{equation}
should be used whenever there is significant variation in multiset size $n$, in which case $n$ can be passed separately.

%% file: sections/relationship.tex
\section{Type relations} \label{sec:typeRelations}

\subsection{Product and tensor} \label{sec:prod&tensor}
A product of tensors can be seen as a refinement of a tensor of a higher order:
\begin{equation}\label{eq:prod&tensor}
    \Prod{\Tensor{l_1, ..., l_{n}}, \, ..., \,\Tensor{l_1, ..., l_{n}}} \ \subtype\ \Tensor{l',l_1, ..., l_{n}}
\end{equation}
where there are $l'$ tensor types in the product type.

For simplicity's sake, we shall investigate the details of this relationship only for the special case where $n=1$:
\begin{equation}
    \Prod{\Vector{l}, \,...,\, \Vector{l}} \subtype\ \Tensor{l', l}  
\end{equation}
Let us convert a vector tuple to a tensor:
\begin{align*}
    \left(\boldsymbol{v}^{(1)}, \,..., \, \boldsymbol{v}^{(l')}\right) &\longrightarrow \; \boldsymbol{N}\;  \\
     v^{(i)}_j \; &\longrightarrow \;  N_{ij} , \quad i=1, ..., l', \quad j=1, ..., l 
\end{align*}
The tensor MFL \eqref{eq:encodingTensor} for order-2 tensors is:
\begin{equation}\label{eq:tensor2MFL}
    y_k = b_k + \sum_{i= 1}^{l'} \sum_{j= 1}^{l}  \left(L^{(1)}_{k\,j} w^{(2)}_{i} + w^{(1)}_{j} L^{(2)}_{k\, i} \right) N_{i j}
\end{equation}
It can be seen as a special case of the product type MFL \eqref{eq:encodingProduct} that is truncated to linear order (i.e.\ higher-order weight tensors are set to zero), with weight matrices constrained to a specific structure:
\begin{equation}
    y_k = b_k + \sum_{i=1}^{l'} \sum_{j=1}^l \tilde{L}_{k\, j}^{(i)} \, v_{j}^{(i)}, \quad 
    \tilde{L}_{k\, j}^{(i)} = L^{(1)}_{k\,j} w^{(2)}_{i} + w^{(1)}_{j} L^{(2)}_{k\, i}
\end{equation}

In conclusion, converting a product type to a tensor type involves discarding the non-linear interaction terms and having a restricted class of weight matrices.

\subsection{Tensor and multiset}
A tensor can be seen as a refined form of a multiset of lower-order tensors:
\begin{equation}
    \Tensor{l', l_1, ..., l_{n}}\ \subtype\ \MSet{\Tensor{l_1, ..., l_{n}}} 
\end{equation}

Like in the previous subsection, we shall investigate the details of this relationship only for the special case where $n=1$:
\begin{equation}
    \Tensor{l', l} \subtype\ \MSet{\Vector{l}}
\end{equation}
Let us convert an order-2 tensor to a multiset of vectors:
\begin{align*}
    \boldsymbol{N}\; &\longrightarrow \; \{ \boldsymbol{v}^{(1)}, ...,  \boldsymbol{v}^{(l')}\}\\
    N_{ij} \;&\longrightarrow \; v^{(i)}_j, \quad i=1, ..., l', \quad j=1, ..., l.   
\end{align*}
The corresponding multiset MFL \eqref{eq:encodingMSet}:
\begin{equation}
    y_k = l'\, b'_k + \sum_{i=1}^{l'} \sum_{j=1}^l L_{k\, j} \, v_{j}^{(i)} 
\end{equation}
is a special case of the tensor MFL \eqref{eq:tensor2MFL}:
\begin{equation}
    y_k = b_k + \sum_{i= 1}^{l'} \sum_{j= 1}^{l}  \left(L^{(1)}_{k\,j} w^{(2)}_{i} + w^{(1)}_{j} L^{(2)}_{k\, i} \right) N_{i j}.
\end{equation}
when:
\begin{equation}
   b_k = l'\, b'_k \quad \text{and} \quad L^{(1)}_{k\,j} w^{(2)}_{i} + w^{(1)}_{j} L^{(2)}_{k\, i} = L_{k\, j}, \quad \forall i
\end{equation}

In conclusion, converting a tensor type to a multiset type results in the weight matrix losing its dependency on the replaced tensor axis.

\subsection{Product, tensor and multiset}
Given the subtyping relationships discussed in the preceding subsections, we conclude that:
\begin{equation*}
    \Prod{\Tensor{l_1, ..., l_n},\, ...,\, \Tensor{l_1,..., l_n}} \ \subtype\ \Tensor{l',l_1,..., l_n}\ \subtype\ \MSet{\Tensor{l_1,..., l_n}}
\end{equation*}
where there are $l'$ tensor types in the product type.

\subsection{Product and multiset}
A generic product type can also be viewed as a refinement of a multiset type:
\begin{equation} \label{eq:prodMSet}
    \Prod{T_1, ..., T_n}\ \subtype\ \Prod{U, ..., U}\ \subtype\  \MSet{U}
\end{equation}
where $U = \texttt{Union}[\,T_1, ...,T_n\,]$.
In converting a product type to a multiset type, some information is lost, such as type tags (or positions).

The MFL for $\MSet{\Vector{l}}$ \eqref{eq:encodingMSet}:
\begin{equation}
    \boldsymbol{y} = n \,\boldsymbol{b} + \sum_{r=1}^n  \boldsymbol{L} \, \boldsymbol{v}^{(r)}
\end{equation}
is a special case of the MFL for $\Prod{\Vector{l}, \,...,\, \Vector{l}}$ \eqref{eq:productMFL_matrixForm}, truncated at the linear order:
\begin{equation}
    \boldsymbol{y} = \boldsymbol{b}' + \sum_{r=1}^{n}  \boldsymbol{L}^{(r)} \, \boldsymbol{v}^{(r)}
\end{equation}
and with the constraints:
\begin{equation}
    \boldsymbol{b}' = n\, \boldsymbol{b} 
    \quad \text{and} \quad  \boldsymbol{L}^{(r)} = \boldsymbol{L}, \quad\forall \,r
\end{equation}

%% file: sections/additional.tex
\section{Additional considerations}\label{sec:additional}
\subsection{Simplification}\label{sec:simplification}

The resulting composed architecture might contain some redundant parts, namely consecutive linear layers. These may be merged into single linear layers. This is usually advantageous, as it can reduce the number of weights and matrix multiplications in the architecture. For example:
\begin{equation}
    \begin{tikzcd}
    V^{(1)} \ar[r,"\boldsymbol{M}"] \ar[rr,bend right,"\boldsymbol{L}"'] & V^{(2)} \ar[r,"\boldsymbol{N}"] & V^{(3)}
\end{tikzcd}
\end{equation}
That is, the two separate linear layers $\boldsymbol{M}$ and $\boldsymbol{N}$ can be replaced with $\boldsymbol{L}$.

\subsection{Weight sharing}

Whenever data transiting within one part of a model shares the exact same type (including semantics) with data in another part, the same submodel should be used for both. Therefore, the type tree—as well as the corresponding architecture—should be reduced to a directed acyclic graph.

Similarly, when data types $T_1$ and $T_2$ share a non-trivial upper bound $T$, a common submodel can manage the shared structure of $T$. However, separate submodels should handle the unique characteristics of $T_1$ and $T_2$.

More generally, if one type can be derived from another, weight sharing is possible. For instance, if $T$ can be projected onto $T'$ in a semantically meaningful way, then a model for $T'$ can be reused as part of a model for $T$.

\subsection{Number types}

A whole paper could be devoted to the encoding of various types of numbers for different contexts. For instance, numbers representing magnitudes should be encoded differently from whole numbers where factorization is significant.

In general, as much information as possible about the number type—such as its structure, semantics, distribution, and usage—should inform the encoding process.

It is often helpful to encode a single number as a coordinate vector, with each component encoding the number (or part of it) in a distinct way.

%% file: sections/example.tex
\section{Examples} \label{sec:example}

\input{sections/featuresExample}

\subsection{Lists}
The list data type is an algebraic data type that is defined recursively:
\begin{equation}
\List{T} = \Sum{\texttt{Nil},\, \Prod{T,\, \List{T}}}
\end{equation}
where the empty list \texttt{Nil} is isomorphic to \Unit.
We expect the flattening architecture of the list type to reflect its recursive structure.

\begin{figure}
    \centering
    \includegraphics[width=0.9\linewidth]{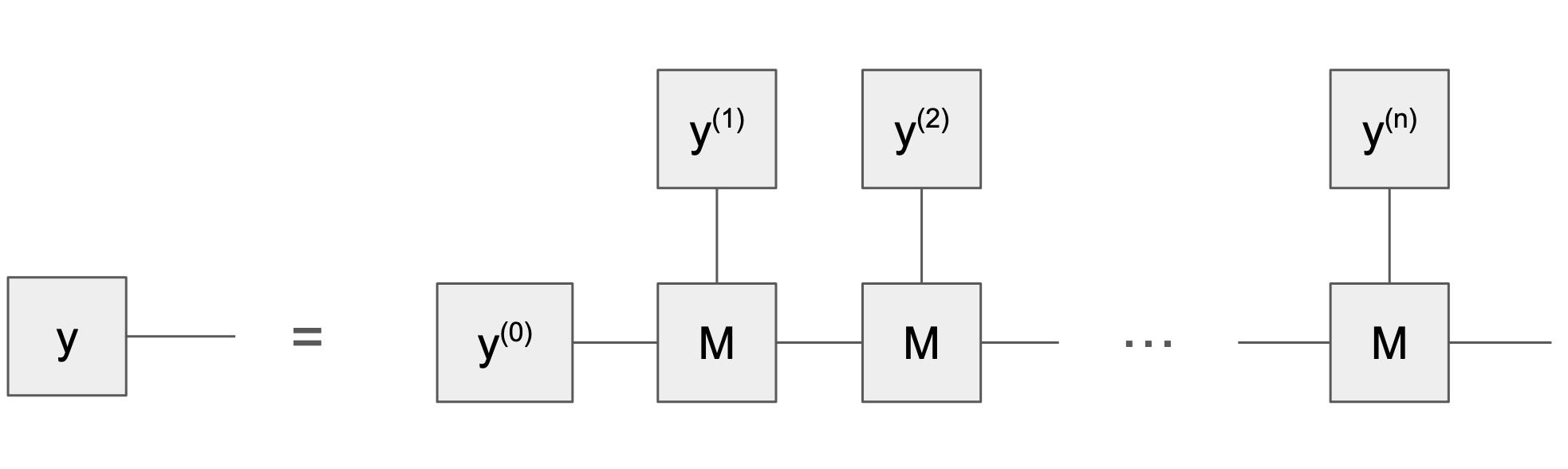}
    \caption{The tensor network diagram of the flattening architecture for list types \eqref{eq:flatteningList}. Shapes are tensors, lines represent indices, and connected lines are contracted indices.}
    \label{fig:tensorDiagramForList}
\end{figure}

\begin{figure}
    \centering
    \includegraphics[width=0.7\linewidth]{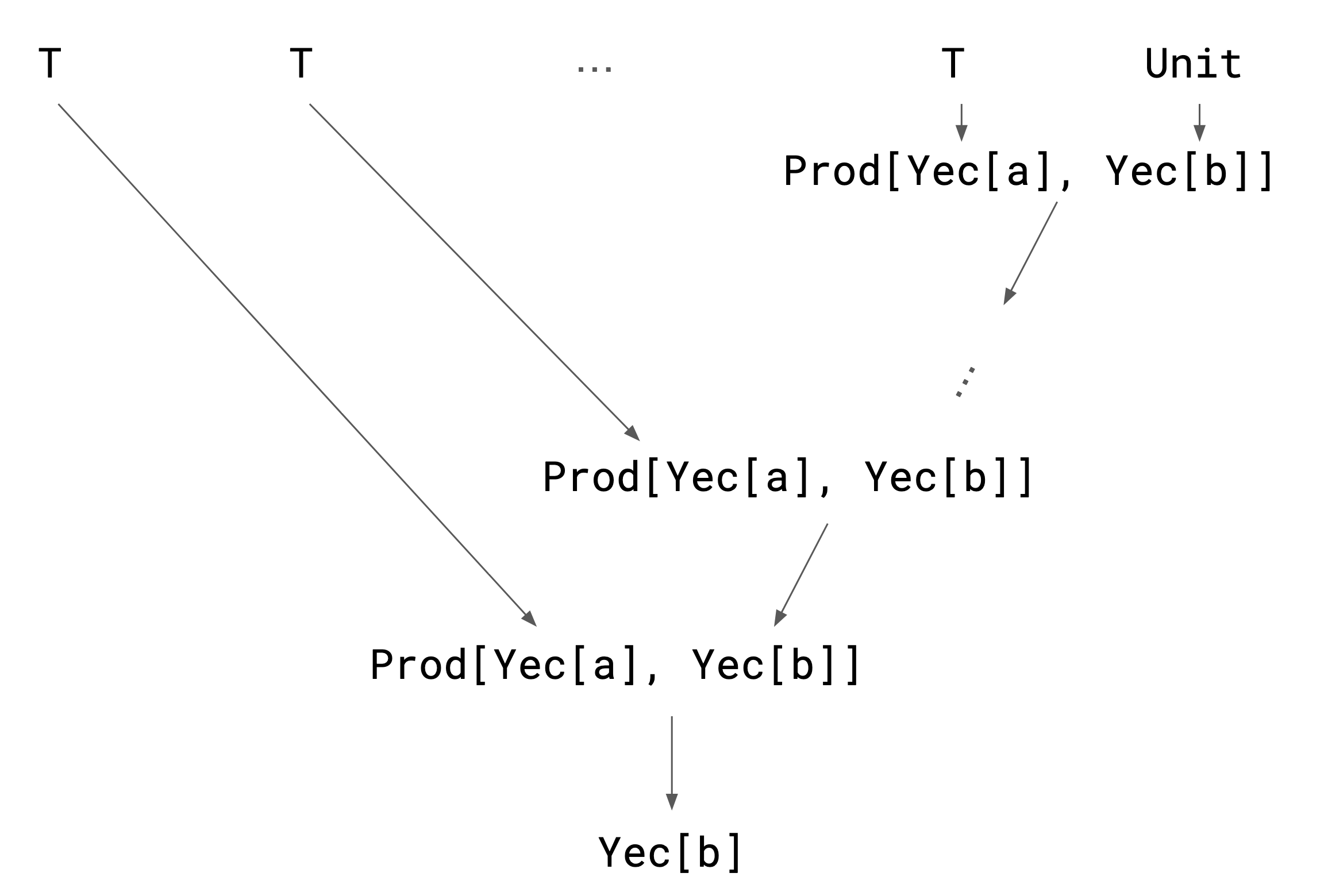}
    \caption{The type tree of the flattening architecture for the list type $\List{T}$, where the empty list (type) \texttt{Nil} is identified with \Unit.}
    \label{fig:typeTreeList}
\end{figure}

Let us consider a list value, built with the (infix) list constructor: 
\begin{equation}
    x_n :: \cdots :: x_1 :: \texttt{Nil} 
\end{equation}
Now, assume we have already flattened \texttt{Nil} (via an embedding) and the elements of the list: 
\begin{align*}
    \texttt{Nil}  &\rightarrow \boldsymbol{y}^{(0)}\\
     x_i &\rightarrow \boldsymbol{y}^{(i)}, \quad i= 1, \ldots, n
\end{align*}
The flattening architecture (in bias-augmented form) is then a recurrent neural network (RNN), with recurrence relation: 
\begin{align}\label{eq:flatteningList}
    \boldsymbol{h}^{(0)}  &= \boldsymbol{y}^{(0)} \nonumber\\
    \boldsymbol{h}^{(i)}  &= \boldsymbol{M}(\boldsymbol{y}^{(i)}, \boldsymbol{h}^{(i-1)}), \quad i=1,\ldots, n \nonumber \\
    \boldsymbol{y}  &=  \boldsymbol{h}^{(n)} 
\end{align}
where $\boldsymbol{M}$ is the bilinear map in the multilinear form of the product type MFL \eqref{eq:encodingProductMultilinear}. The unfolded version is depicted in fig.~\ref{fig:tensorDiagramForList} using a tensor network diagram. The type tree is shown in fig.~\ref{fig:typeTreeList}. Of course, non-linear activation functions can be applied to the hidden states $\boldsymbol{h}^{(i)}$.

%% file: sections/featuresExample.tex
\subsection{Features}

\begin{figure}
    \begin{center}
         \includegraphics[width=0.7\linewidth]{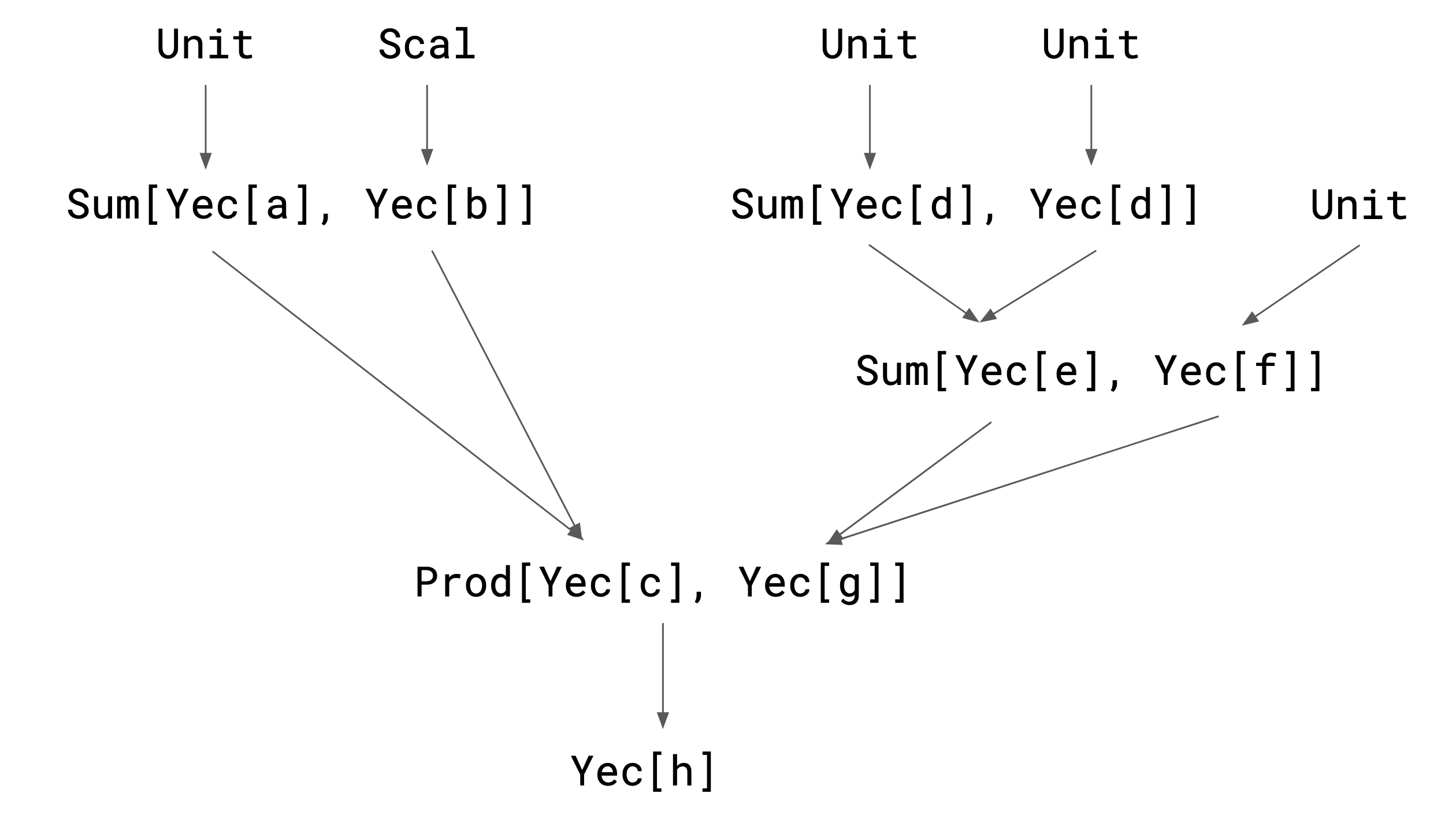}
    \end{center}
    \caption{The type tree of the flattening of the product type \eqref{eq:example1}. We used the type synonyms: $\Option{\Scalar} = \Sum{\Unit,\, \Scalar}$ and $\Bool = \Sum{\Unit,\, \Unit}$.}
    \label{fig:example1}
\end{figure}

\begin{figure}
    \centering
    \includegraphics[width=0.65\linewidth]{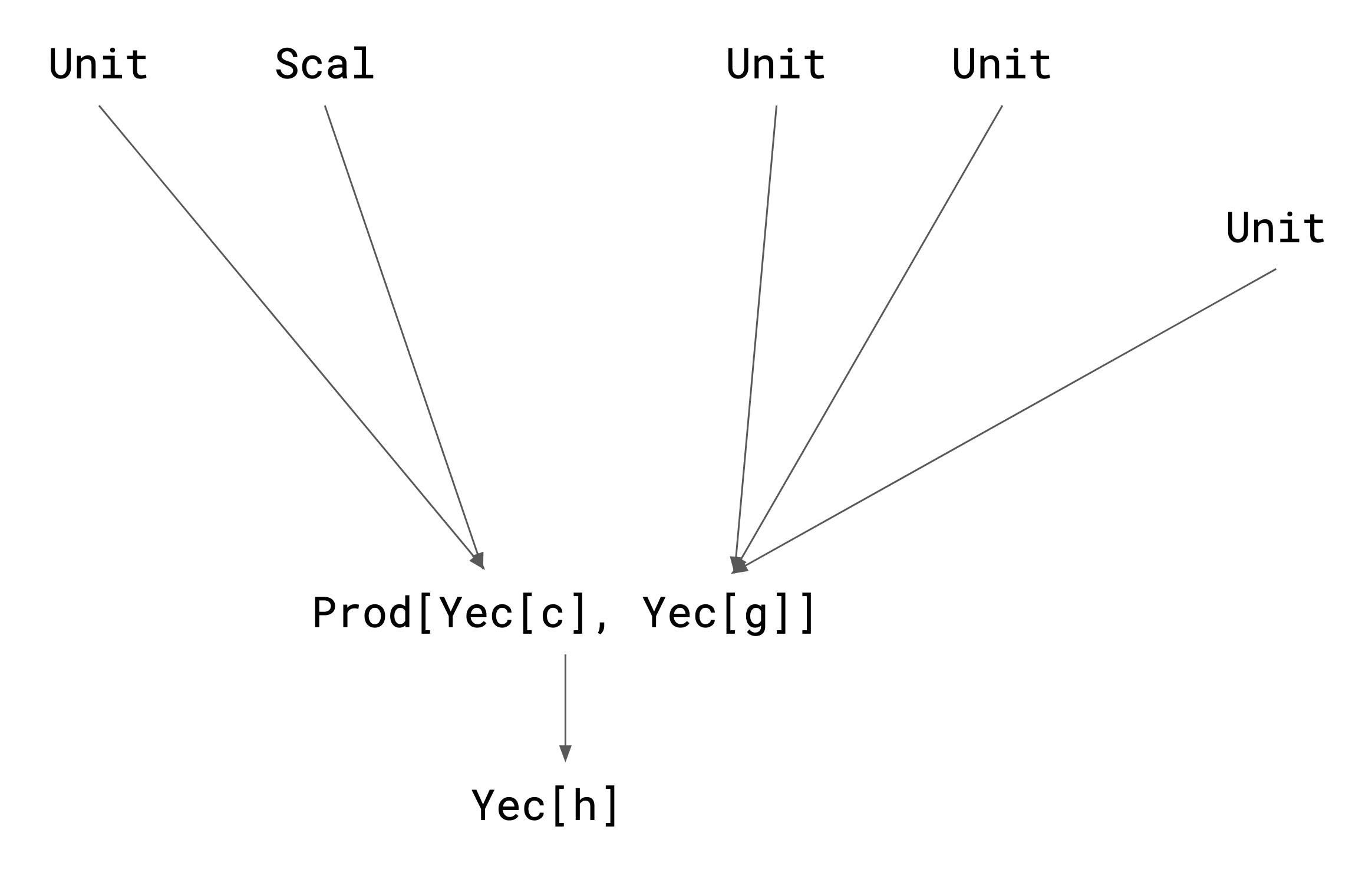}
    \caption{A simplified type tree of the flattening of \eqref{eq:example1}.} 
    \label{fig:simplifiedExample1}
\end{figure}

Consider two input features with the following types:
\begin{itemize}
    \item $\Option{\Scalar}$
    \item $\Option{\Bool}$
\end{itemize}
where the option type takes into account that some values may be missing in the training data. These features are combined into a product type:
\begin{eqnarray}\label{eq:example1}
    \Prod{\Option{\Scalar}, \, \Option{\Bool}}
\end{eqnarray}
The flattening architecture for this composite type is shown in fig.~\ref{fig:example1}. Let us consider the case where only MFLs are used. Then, the flattening architecture for the product type above can be further simplified using linear composition, see fig.~\ref{fig:simplifiedExample1}, as discussed in section \ref{sec:simplification}. The final flattened yector is:
\begin{equation}
    \boldsymbol{y} = \boldsymbol{b} + \boldsymbol{L}^{(1)} \boldsymbol{y}^{(1)} + \boldsymbol{L}^{(2)} \boldsymbol{y}^{(2)}  +  \boldsymbol{C}^{(1,2)}( \boldsymbol{y}^{(1)}, \boldsymbol{y}^{(2)} )
\end{equation}
where $\boldsymbol{b}$ is the bias vector, $\boldsymbol{L}^{(1)}$ and $\boldsymbol{L}^{(1)}$ are weight matrices, $\boldsymbol{C}^{(1,2)}$ is a rank-3 weight tensor, and:
\begin{align*}
    \boldsymbol{y}^{(1)} &= 
    \begin{cases}
      \; \boldsymbol{b}^{(1)} & \text{for} \quad \Unit \: \text{(i.e. missing value)}\\
      \;\boldsymbol{b}^{(2)} + s \, \boldsymbol{w}  & \text{for}\quad  \Scalar 
    \end{cases}  \\
    \boldsymbol{y}^{(2)} &= 
    \begin{cases}
      \;\boldsymbol{b}^{(3)} & \qquad \;\; \,\text{for} \quad \Unit_1 \: \text{(i.e. true)}\\
      \;\boldsymbol{b}^{(4)}  & \qquad \;\; \,\text{for}\quad \Unit_2 \: \text{(i.e. false)}\\
      \;\boldsymbol{b}^{(5)}  & \qquad \;\; \,\text{for}\quad \Unit_3 \: \text{(i.e. missing value)}\\      
    \end{cases}      
\end{align*}
where we added subscripts to show the position of the unit types in the sum type.

Note that the flattening architecture for the product type automatically takes into account the correlation between inputs.

%% file: sections/conclusion.tex
\section{Conclusion}\label{sec:conclusion}

In this paper, we introduced the primitives for an algebra allowing the construction of model architectures that respect the structure of types constructed inductively from sums, products, and multisets, with the tensor type as the base case. Together, we refer to these types as algebraic data types (ADTs).

We characterized each ADT by a core set of primitive operations and explored the relationship between tensors, product types and multisets, which are sometimes conflated in machine learning practice.  

Our key contribution is multilinear flattening layers (MFLs), which generalize dense linear layers to accommodate ADTs.
Notably, the MFL for tensor types \eqref{eq:encodingTensor} naturally exhibits a low-rank decomposition due to the inherent structure of tensors, whereas the MFL for product types \eqref{eq:encodingProduct} is full-rank.

By way of illustration, we presented minimal flattening architectures for a product of option types, and for the recursively defined type $\List{T}$. The latter naturally results in a recurrent architecture. In a similar fashion, architectures for trees and graphs can be derived.

Follow-up work could focus on the proper handling of additional factors that should influence architecture, such as:
\begin{itemize}
    \item Input structure (including abstract data types)
    \item Output structure
    \item Implicit structure preservation
    \item Algorithmic/compositional structure
\end{itemize}

The ultimate goal is a solid compositional foundation for structured and typeful machine learning.

%% file: sections/appendix.tex
\section{Appendix}

Let us explicitly show the terms containing the 0th dimension in the following sum: 
\begin{align} \label{eq:multiSumWithBias}
    \sum_{i_1, ..., i_n} M_{i_1\cdots i_n }  
    &=M_{0\cdots 0 }  \nonumber\\
     &+ \sum_{j_1}  M_{j_1 0\cdots 0 }  + \cdots 
     + \sum_{j_n}  M_{ 0\cdots 0 \,j_n }  \nonumber\\
    &+  \sum_{j_1, j_2}  M_{j_1 j_2 0\cdots 0}   + \cdots 
    + \sum_{j_{n-1},j_n}  M_{ 0\cdots 0 \,j_{n-1} j_n} \nonumber\\
    &\;\,\vdots\nonumber\\
    &+ \sum_{j_1, ..., j_{n-1}} M_{ j_1\cdots j_{n-1}0} + \cdots 
    +  \sum_{j_2, ..., j_n}  M_{0 j_2... j_n} \nonumber\\
    &+\sum_{j_1, ..., j_n } M_{j_1... j_n} 
\end{align}
where the $j$ indices start from $1$. 

In the case of the tensor MFL \eqref{eq:augmentedTensorFlattening}, all the sum terms containing the bias components can be collapsed into one single bias vector, leading to \eqref{eq:encodingTensor}.

For the product type MFL, we have:
\begin{align} \label{eq:encodingProductExpanded}
    v_k
    &=M_{k\, 0... 0} \nonumber\\
    &+ \sum_{j_1}  M_{k\,j_1 0\cdots 0 } \, v_{j_1}^{(1)} + \cdots + \sum_{j_n}  M_{k\,0\cdots 0 \,j_n} \, v_{j_n}^{(n)} \nonumber \\
    &+ \sum_{j_1, j_2}  M_{k\,j_1 j_2 0\cdots 0 } \, v_{j_1}^{(1)}v_{j_2}^{(2)} + \cdots + \sum_{j_{n-1},j_n}  M_{k\, 0\cdots 0\, j_{n-1} j_n } \, v_{j_{n-1}}^{(n-1)} v_{j_n}^{(n)}   \nonumber \\
    &\;\,\vdots  \nonumber\\
    &+ \sum_{j_1, ..., j_{n-1}}  M_{k\,j_1\cdots j_{n-1}0 } \, v_{j_1}^{(1)} \cdots v_{j_{n-1}}^{(n-1)}  + \cdots + \sum_{j_2, ..., j_n}  M_{k\,0 j_2\cdots j_n} \, v_{j_2}^{(2)} \cdots v_{j_n}^{(n)}   \nonumber\\    
    &+\sum_{j_1, ..., j_n } M_{k\, j_1\cdots j_n } \, v_{j_1}^{(1)} \cdots v_{j_n}^{(n)}.
\end{align}